\documentclass[11pt,a4paper]{article}
\usepackage[hyperref]{emnlp2020}
\usepackage{times}
\usepackage{latexsym}

\usepackage{microtype}

\usepackage{amsmath,booktabs,multirow,enumerate,subcaption,makecell,graphicx,lipsum}

\aclfinalcopy 

\title{Cost-effective Selection of Pretraining Data:\\A Case Study of Pretraining BERT on Social Media}

\author{\begin{tabular}{cccc}
Xiang Dai$^{1,2}$ & Sarvnaz Karimi$^{1}$ & \textbf{Ben Hachey$^{3}$} & \textbf{Cecile Paris$^{1}$}
\end{tabular}\\
\begin{tabular}{cccc}
\multicolumn{4}{c}{$^{1}$CSIRO Data61, Sydney, Australia}\\
\multicolumn{4}{c}{$^{2}$University of Sydney, Sydney, Australia}\\
\multicolumn{4}{c}{$^{3}$Harrison ai, Sydney, Australia}\\
\multicolumn{4}{c}{\tt \{dai.dai,sarvnaz.karimi,cecile.paris\}@csiro.au}\\
\multicolumn{4}{c}{\tt ben.hachey@gmail.com} \\
\end{tabular}
}

\date{}

\begin{document}
\maketitle
\begin{abstract}

Recent studies on domain-specific BERT models show that effectiveness on downstream tasks can be improved when models are pretrained on in-domain data. Often, the pretraining data used in these models are selected based on their subject matter, e.g., biology or computer science. Given the range of applications using social media text, and its unique language variety, we pretrain two models on tweets and forum text respectively, and empirically demonstrate the effectiveness of these two resources. In addition, we investigate how similarity measures can be used to nominate in-domain pretraining data. We publicly release our pretrained models at \text{https://bit.ly/35RpTf0}.
\end{abstract}

\section{Introduction}
Sequence transfer learning~\citep{Ruder:Transfer:2019}, that pretrains language representations on unlabeled text (\emph{source}) and then adapts these representations to a supervised task (\emph{target}), has demonstrated its effectiveness on a range of NLP  tasks~\citep{Radford:Narasimhan:OpenAI:2018,Devlin:Chang:NAACL:2019,Liu:Ott:arXiv:2019}. Approaches vary in model, pretraining objective, pretraining data and adaptation strategy. We consider a widely used method, BERT~\citep{Devlin:Chang:NAACL:2019}. It pretrains a transformer-based model using a masked language model objective and then fine-tunes the model on the target task. We investigate the impact of the domain (i.e., the similarity between the underlying distribution of source and target data) of pretraining data on the effectiveness of pretrained models. We also propose a cost-effective way to select pretraining data. 

Recent studies on domain-specific BERT models, which are pretrained on specialty source data, empirically show that, when in-domain data is used for pretraining, target task performance can be  improved~\citep{Lee:Yoon:arXiv:2019,Alsentzer:Murphy:ClinicalNLP:2019,Huang:Altosaar:arXiv:2019,Beltagy:Lo:EMNLP:2019}. These publicly available domain-specific BERT models are valuable to the NLP community. 
However, the selection of in-domain data usually resorts to intuition, which varies across NLP practitioners~\citep{Dai:Karimi:NAACL:2019}. According to~\citet{Halliday:Hasan:1989}, the context specific usage of language is affected by three factors: \emph{field} (the subject matter being discussed), \emph{tenor} (the relationship between the participants in the discourse and their purpose) and \emph{mode} (communication medium, e.g., `spoken’ or `written’).\footnote{We do not explicitly consider mode in this study, because all data used are written text.} Generally, the selection of pretraining data in existing domain-specific BERT models is based on the field rather than the tenor. For example, BioBERT~\citep{Lee:Yoon:arXiv:2019} and SciBERT~\citep{Beltagy:Lo:EMNLP:2019} are both pretrained on scholar articles, but on different fields (biology and computer science).

\begin{figure*}
    \centering
    \includegraphics[width=0.85\textwidth]{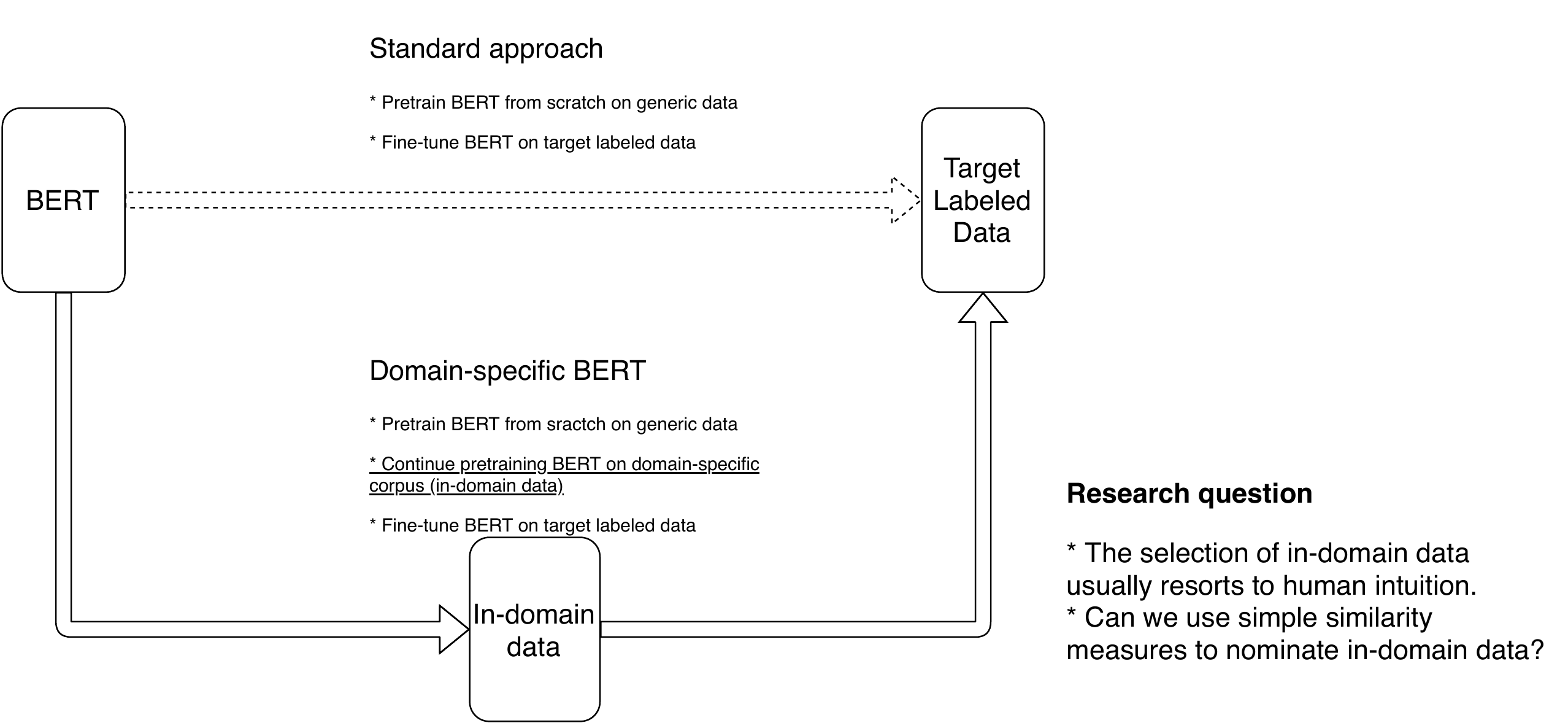}
    \caption{Recent studies have demonstrated the effectiveness of domain-specific BERT models. However, the selection of in-domain data usually resorts to intuition, which varies across NLP practitioners, especially regarding intersecting domains. We investigate the correlation of source-target similarity and the effectiveness of pretrained models. In other words, we aim to use simple similarity measures to nominate in-domain pretraining data.~\label{figure-question}}
\end{figure*}

We conduct a case study of pretraining BERT on social media text which has very different tenor from existing domain-specific BERT models.
Our contributions are two-fold: (1) We release two pretrained BERT models trained on tweets and forum text, and we demonstrate the effectiveness of these two resources on a range of NLP data sets using social media text; and, (2) we investigate the correlation of source-target similarity and task accuracy using different domain-specific BERT models. We find that simple similarity measures can be used to nominate in-domain pretraining data (Figure~\ref{figure-question}).
    
\section{Related Work}
\paragraph{Selecting data to pretrain BERT}
There are two known strategies: (1) collecting very large  generic data, such as web crawl and news~\citep{Radford:Wu:OpenAI:2019,Liu:Ott:arXiv:2019,Baevski:Edunov:arXiv:2019}; and, (2) selecting in-domain data, which we refer to as domain-specific BERT models.

Those following the first strategy intend to build universal language representations that are useful across multiple domains. They also believe that pretraining on larger data leads to better pretrained models. For example, \citet{Baevski:Edunov:arXiv:2019} empirically show that the average GLUE score~\citep{Wang:Singh:ICLR:2019} can increase from lower than 80 to higher than 81 when the size of pretraining data increases from 562 million to 18 billion tokens.

Our study uses the second strategy. However, we select our pretraining data from the tenor perspective rather than the field. A summary of the source data used in these domain-specific BERT models can be found in Table~\ref{table-comparision-source-data}.

\paragraph{Finding in-domain data}
Our study relates to the literature on investigating domain similarity~\citep{Blitzer:McDonald:EMNLP:2006,Ben-David:Blitzer:NIPS:2007,Ruder:Plank:EMNLP:2017} and text similarity~\citep{Mihalcea:Corley:AAAI:2006,Pavlick:Rastogi:ACL:2015,Kusner:Sun:ICML:2015}. Our work is also inspired by the study by~\citet{Dai:Karimi:NAACL:2019} on the impact of source data on pretrained LSTM-based models (i.e., ELMo) and by~\citet{Asch:Daelemans:DANLP:2010} on the correlation between similarity and accuracy loss of POS taggers.

\begin{table}[t]
    \small
    \centering
    \setlength{\tabcolsep}{2.5pt} 
    \begin{tabular}{p{0.4\linewidth}p{0.55\linewidth}}
    \toprule
    \bf Model & \bf Source data \\
    \midrule
    Original BERT & Books and encyclopedia articles, various fields \\
    BioBERT~\cite{Lee:Yoon:arXiv:2019} & Scholar articles on biology \\
    ClinicalBERT~\cite{Alsentzer:Murphy:ClinicalNLP:2019} & Nursing and physician notes on hospital admission \\
    SciBERT~\cite{Beltagy:Lo:EMNLP:2019} & Scholar articles on biology and computer science \\
    TwitterBERT (this work) & Tweets, various fields \\
    ForumBERT (this work) & Forum text on business review \\
    \bottomrule
    \end{tabular}
    \caption{A summary of source data used in the original BERT and several domain-specific BERT models.\label{table-comparision-source-data}}
\end{table}

\section{Pretraining BERT Models}

We follow the practices used in other domain-specific BERT models~\citep{Lee:Yoon:arXiv:2019,Beltagy:Lo:EMNLP:2019} to pretrain our BERT models. We use the original vocabulary of BERT-Base as our underlying word piece vocabulary\footnote{\citet{Beltagy:Lo:EMNLP:2019} investigated the effect of having an in-domain vocabulary. Their results show that, although an in-domain vocabulary is helpful, the magnitude of improvement is relatively small.} and use the pretrained weights from the original BERT-Base as the initialization weights. Note that all domain-specific models we consider in this study are based on this paradigm,\footnote{We notice a very recent resource by~\citet{Nguyen:Vu:arXiv:2020} who pretrain RoBERTa on general English tweets, as well as tweets related to the COVID-19 pandemic. We did not consider this model as it involves more variants: byte pair encoding and initialization weights.} which means these models are supposed to capture both generic (inheriting from original BERT) and domain-specific knowledge.

For pretraining objective, we remove the Next Sentence Prediction (NSP) objective. Social media text, especially tweets, are often too short to sample consecutive sentences. In addition, recent studies observe benefits in removing the NSP objective with sequence-pair training~\citep{Liu:Ott:arXiv:2019}.

\paragraph{Twitter}
We use English tweets ranging from Sep 1 to Oct 30, 2018\footnote{\href{https://archive.org/details/archiveteam-twitter-stream-2018-09}{Internet archive}, Accessed 1 June 2020.} to pretrain our Twitter BERT. There are in total 60 million English tweets, consisting of $0.9$B tokens. Although we aim to avoid tailored pre-processing strategies to make a fair comparison with other domain-specific BERT models, we find 44\% of these tweets contain url and 78\% contain other user names (@, if a tweet replies another tweet, @ is added automatically). We thus employ minimal processing by: (1) replacing tokens starting with `@', referring to a Twitter user's account name, with a special token [TwitterUser]; and, (2) replacing urls as a special token [URL]. We hypothesize that the surface form of these tokens do not contain useful information.

\paragraph{Forum}
We use local businesses reviews released by Yelp\footnote{\href{https://www.yelp.com/dataset}{Yelp Challenge}, Accessed 1 June 2020.} to pretrain our Forum BERT. There are in total five million reviews, consisting of $0.6$B tokens. No preprocessing is conducted on the text.

We used four Nvidia P100 GPUs for the pretraining. Training of each model took seven days.

\section{Effectiveness of Pretrained BERT Models}
To evaluate the effectiveness of our pretrained BERT models, we experiment on a range of classification and Named Entity Recognition (NER) data sets. 
Both text classification and NER are fundamental NLP tasks that can employ generic architectures on top of BERT. For the classification task, the representation of the first token (i.e., [CLS]) is fed into the output layer for the final prediction. 
For the NER task, the representations of the first sub-token within each token are taken as input to a token-level classifier to predict the token's tag.
We did not explore more complex architectures, such as adding LSTM or CRF on top of BERT~\citep{Beltagy:Lo:EMNLP:2019,Baevski:Edunov:arXiv:2019}, because our aim is to demonstrate the efficacy of domain-specific BERT models and to observe the impact of pretraining data, rather than to achieve state-of-the-art performance on these data sets.

Our BERT results follow the standard two-stage approach of finetuning the pretrained model. Domain-specific BERTs add a stage in the middle: finetuning BERT on domain-specific unlabeled data (cf. Figure~\ref{figure-question}).

\subsection{Target Tasks}

\begin{table*}[ht]
    \small
    \centering
    \begin{tabular}{c|l|c|c|c|c|c|c}
    \toprule
    \bf Target Text type& \bf Corpus & \bf BERT & \bf Bio & \bf Clinical & \bf Sci & \bf Twitter & \bf Forum \\
    & & (3.3B) & (18B) & (0.5B) & (3.1B) & (0.9B) & (0.6B) \\
    \midrule
    \multirow{4}{*}{Tweets} & Airline ($C$) & 80.5\scriptsize{$\pm$ 0.3} & 79.0\scriptsize{$\pm$ 0.5} & 78.8\scriptsize{$\pm$ 0.8} & 78.8\scriptsize{$\pm$ 0.9} & 80.8\scriptsize{$\pm$ 0.6} & \underline{\bf 81.6\scriptsize{$\pm$ 0.5}} \\ 
    & BTC ($N$) & 78.0\scriptsize{$\pm$ 0.5} & 75.2\scriptsize{$\pm$ 0.3} & 76.9\scriptsize{$\pm$ 0.5} & 77.4\scriptsize{$\pm$ 0.4} & \bf 79.0\scriptsize{$\pm$ 0.5} & 77.0\scriptsize{$\pm$ 0.4} \\ 
    & SMM4H-18 task3 ($C$) & 76.5\scriptsize{$\pm$ 0.9} & 75.4\scriptsize{$\pm$ 1.1} & 75.6\scriptsize{$\pm$ 0.7} & 75.4\scriptsize{$\pm$ 1.0} & 77.0\scriptsize{$\pm$ 1.0} & \bf 77.2\scriptsize{$\pm$ 1.3}  \\ 
    & SMM4H-18 task4 ($C$) & 89.4\scriptsize{$\pm$ 0.5} & 87.7\scriptsize{$\pm$ 0.4} & 88.1\scriptsize{$\pm$ 0.8} & 88.7\scriptsize{$\pm$ 0.8} & 90.3\scriptsize{$\pm$ 0.3} & \underline{\bf 91.1\scriptsize{$\pm$ 0.6}}  \\ 
    \hline
    \multirow{4}{*}{Forum} & CADEC ($N$) & 71.9\scriptsize{$\pm$ 0.6} & 72.1\scriptsize{$\pm$ 0.6} & 72.1\scriptsize{$\pm$ 0.8} & \underline{\bf 73.2\scriptsize{$\pm$ 0.4}} & 72.1\scriptsize{$\pm$ 1.0} & 72.9\scriptsize{$\pm$ 0.6} \\ 
    & SemEval-14 laptop ($N$) & 81.1\scriptsize{$\pm$ 0.8} & 79.3\scriptsize{$\pm$ 0.3} & 78.5\scriptsize{$\pm$ 0.4} & \bf 81.6\scriptsize{$\pm$ 1.1} & 81.3\scriptsize{$\pm$ 0.6} & 81.4\scriptsize{$\pm$ 1.1} \\ 
    & SemEval-14 restaurant ($N$) & 87.5\scriptsize{$\pm$ 0.6} & 84.9\scriptsize{$\pm$ 0.3} & 85.5\scriptsize{$\pm$ 0.7} & 86.7\scriptsize{$\pm$ 0.5} & 87.4\scriptsize{$\pm$ 0.7} & \underline{\bf 89.3\scriptsize{$\pm$ 0.5}} \\ 
    & SST-2 ($C$) & 92.4\scriptsize{$\pm$ 0.2} & 91.1\scriptsize{$\pm$ 0.5} & 90.4\scriptsize{$\pm$ 0.3} & 91.4\scriptsize{$\pm$ 0.4} & 92.3\scriptsize{$\pm$ 0.4} & \underline{\bf 93.4\scriptsize{$\pm$ 0.4}} \\ 
    \hline
    \multirow{3}{*}{Non-social media} & EBM ($N$) & 41.5\scriptsize{$\pm$ 0.5} & 42.1\scriptsize{$\pm$ 0.2} & 41.1\scriptsize{$\pm$ 0.5} & \bf 42.4\scriptsize{$\pm$ 0.7} & 40.5\scriptsize{$\pm$ 0.5} & 41.5\scriptsize{$\pm$ 0.5} \\ 
    & i2b2-10 ($N$) & 85.8\scriptsize{$\pm$ 0.1} & \underline{\bf 87.4\scriptsize{$\pm$ 0.2}} & \underline{\bf 87.4\scriptsize{$\pm$ 0.1}} & 87.3\scriptsize{$\pm$ 0.2} & 84.8\scriptsize{$\pm$ 0.2} & 85.2\scriptsize{$\pm$ 0.1} \\ 
    & JNLPBA ($N$) & 72.5\scriptsize{$\pm$ 0.3} & \underline{\bf 74.2\scriptsize{$\pm$ 0.2}} & 71.9\scriptsize{$\pm$ 0.1} & 73.6\scriptsize{$\pm$ 0.3} & 72.2\scriptsize{$\pm$ 0.2} & 72.5\scriptsize{$\pm$ 0.2} \\ 
    & Paper Field ($C$) & 74.5\scriptsize{$\pm$ 0.1} & 74.3\scriptsize{$\pm$ 0.1} & 73.3\scriptsize{$\pm$ 0.1} & \underline{\bf 75.1\scriptsize{$\pm$ 0.1}} & 74.1\scriptsize{$\pm$ 0.1} & 73.3\scriptsize{$\pm$ 0.2} \\
    \bottomrule
    \end{tabular}
    \caption{Effectiveness of different BERT models, evaluated on downstream tasks. \# tokens in each pretraining data are listed in brackets. $C$: Classification task, for which we report macro-F1; $N$: NER task, for which we report span-level micro-F1. We repeat all experiments five times with different random seeds. Mean values are reported. \underline{underline}: the best result is significantly better than the second best result (paired student's t-test, p: $0.05$).\label{table-bert-results}}
\end{table*}


We use eight target tasks with their text sampled from Twitter and forums, to examine whether our BERT models can lead to improvements, compared to the original BERT. These tasks are \textbf{Airline}\footnote{\href{https://www.kaggle.com/crowdflower/twitter-airline-sentiment}{Kaggle Twitter US Airline Sentiment Challenge}}: classifying sentiment on tweets about major U.S. airlines; \textbf{BTC}: identifying location, person, and organization on tweets~\citep{Derczynski:Bontcheva:COLING:2016}; \textbf{SMM4H-18}: classifying whether the user reports an adverse drug events (task3)~\citep{Weissenbacher:Sarker:SMM4H:2018}, or intends to receive a seasonal influenza vaccine (task4) on tweets about health~\citep{Joshi:Dai:SMM4H:2018};
\textbf{CADEC}: identifying adverse drug events etc. on reviews about medications~\citep{Karimi:Metke:JBI:2015}; \textbf{SemEval-14}: identifying product or service attributes on reviews about laptops and restaurants~\citep{Pontiki:Galanis:SemEval:2014}; \textbf{SST}: classifying sentiment on movie reviews~\citep{Socher:Perelygin:EMNLP:2013}.

In addition, we use four tasks that do not use social media text to investigate how our BERT models perform on out-of-domain target tasks: \textbf{Paper Field}: classifying the research topic based on the title of scholar articles about various fields~\citep{Beltagy:Lo:EMNLP:2019}; \textbf{EBM}: identifying intervention, outcome etc. on scholar articles about clinical trials~\citep{Nye:Li:ACL:2018}; \textbf{i2b2-10}: identifying treatment, test and problem on clinical notes about health~\citep{Uzuner:South:AMIA:2011}; \textbf{JNLPBA}: identifying RNA, DNA etc. on scholar articles about biology~\citep{Kim:Ohta:BioNLP:2004}.


\begin{figure}[h!]
  \centering
  \begin{subfigure}[b]{0.9\linewidth}
    \includegraphics[width=\linewidth]{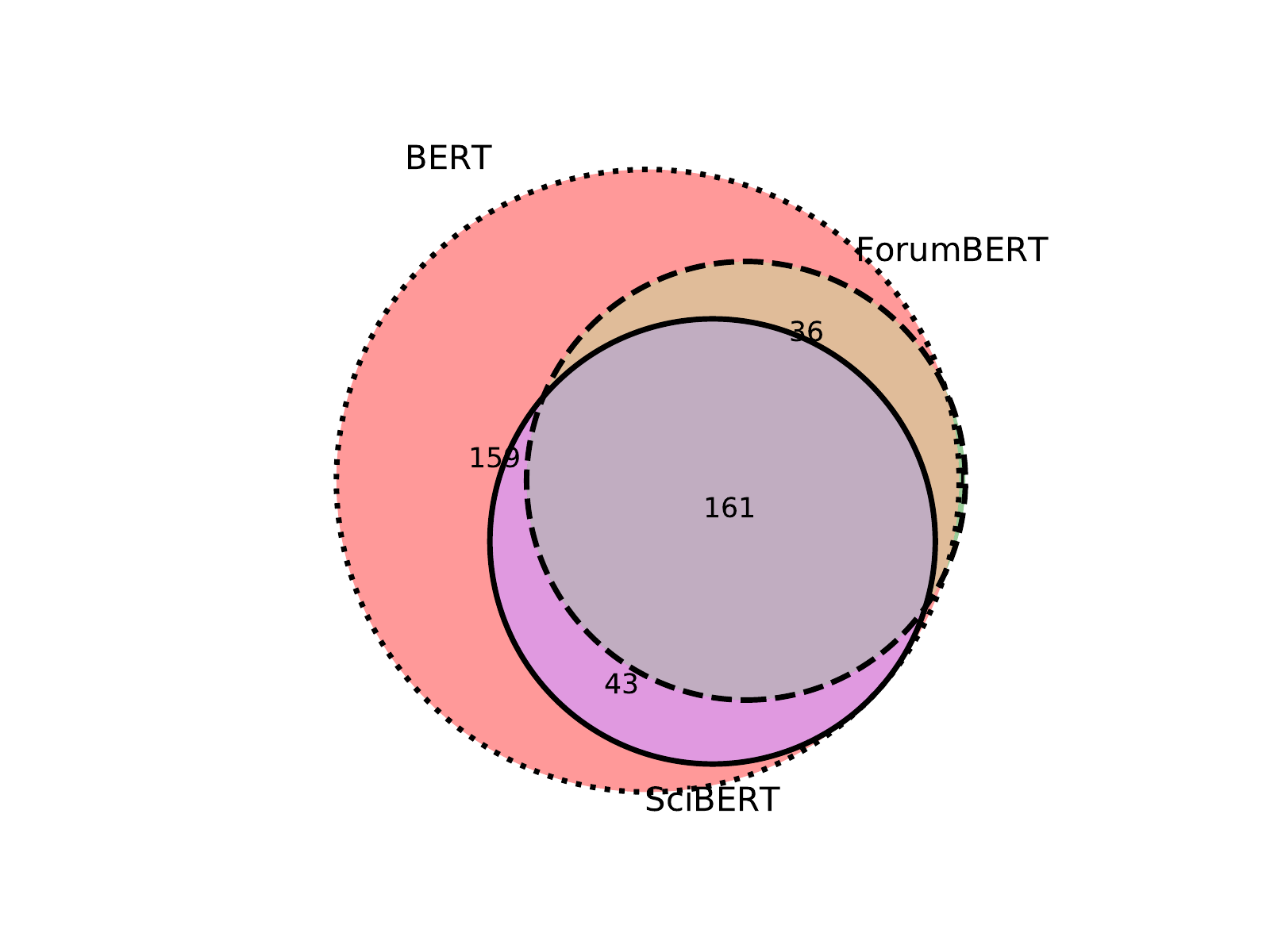}
    \caption{False positives.~\label{figure-error-analysis-fp}}
  \end{subfigure}
  \begin{subfigure}[b]{0.9\linewidth}
    \includegraphics[width=\linewidth]{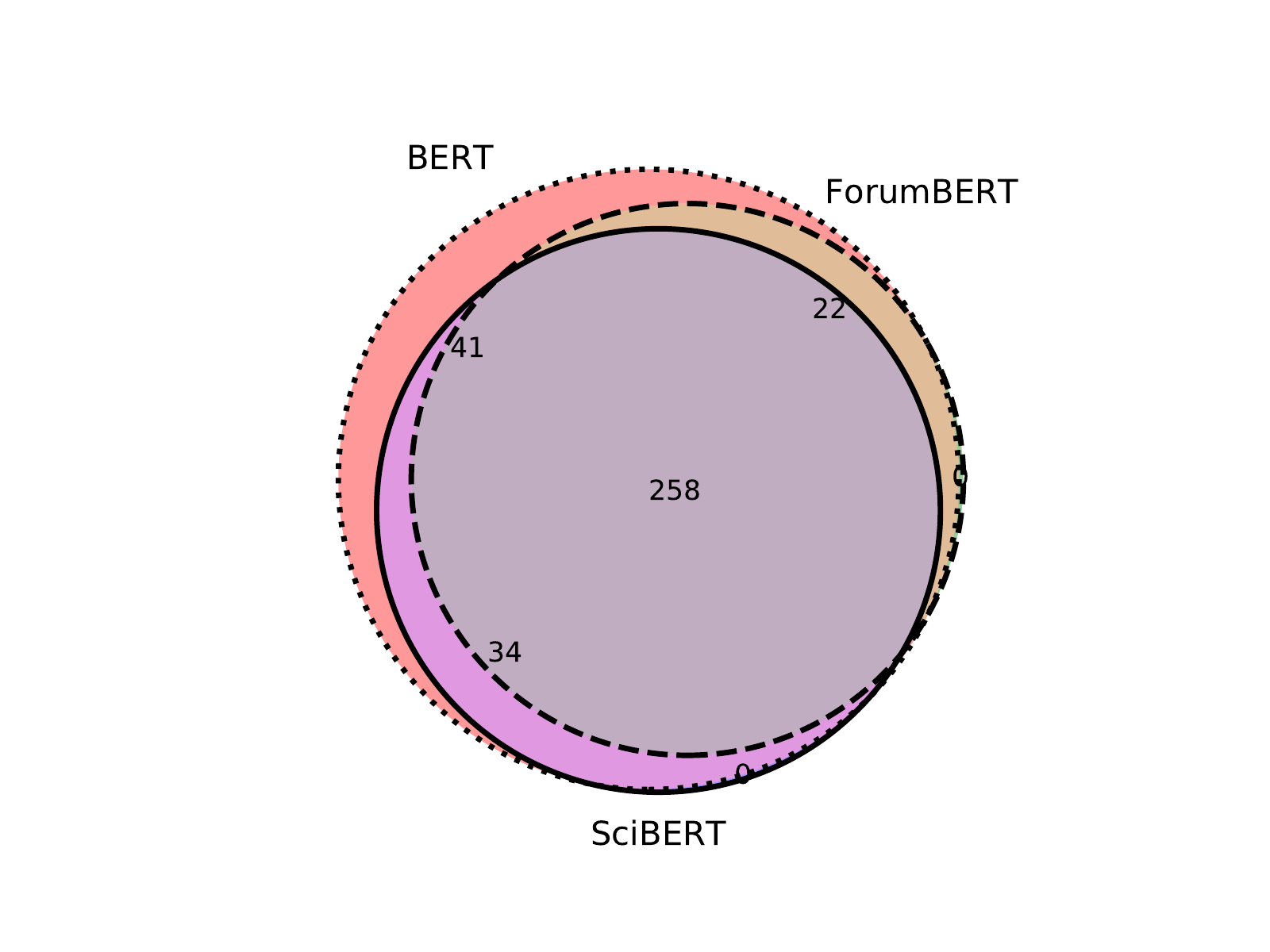}
    \caption{False negatives.~\label{figure-error-analysis-fn}}
  \end{subfigure}
  \caption{Error predictions on CADEC. Dotted line circle: errors by the BERT model. Dashed line: YelpBERT. Solid line: SciBERT.}
\end{figure}

\subsection{Results}

We observe that our BERT models achieve the highest F1 score on 6 out of 8 target tasks that use social media text (Table~\ref{table-bert-results}). On CADEC (medications) and SemEval-14 laptop, SciBERT achieves the highest score due to the overlapping fields (i.e., medication and computer hardware, respectively). We note, however, that our Forum BERT achieves very close results. This demonstrates the effectiveness of our pretrained models on target tasks using social media text. 
To our surprise, on target tasks using tweets, forum BERT achieves better results than Twitter BERT on 3 classification tasks. 
On one hand, this may be explained by~\citet{Baldwin:Cook:IJCNLP:2013}'s observation that forum text is the `median' data, which is similar to all other types of social media text.
On the other hand, it also reveals the challenge of pretraining contextual language representations on short tweets.

We also observe that, when domain-specific models are applied on a target task with out-of-domain data, they achieve much lower results than the original BERT. For example, BioBERT achieves lower results than the original BERT on 7 out of 8 target social media tasks. It only achieves a better result on CADEC, which is about medications. Recall that all these domain-specific BERT models use the pretrained weights of the original BERT as initialization. On one hand, we argue that this observation may challenge the conventional wisdom that the larger the pretraining data is, the better the pretrained model is. Training on out-of-domain source data may cause negative impact, at least for the two-stage pretraining approach we consider. On the other hand, this observation reinforces recent work showing the importance of task-adaptive pretraining~\citep{Gururangan:Marasovic:ACL:2020}.

\paragraph{Error analysis on CADEC}
We conduct an error analysis on CADEC, because it is at the intersection between social media tenor (online posts) and medication field (adverse drug events), and thus could be similar to multiple sources. 
We compare the error predictions by the two best performing BERT models -- ForumBERT and SciBERT, as well as the baseline BERT model.
In Figure~\ref{figure-error-analysis-fp}, we observe that both domain-specific BERT models can reduce greatly the number of false positives made by the baseline BERT. Specifically, 159 false positives made by the baseline BERT are fixed by the domain-specific BERT models. However, domain-specific BERT models do not reduce a lot the number of false negatives -- gold mentions not recognized. There are 258 gold mentions recognized by none of three models, and only 41 false negatives by the baseline BERT are fixed by the domain-specific BERT models (Figure~\ref{figure-error-analysis-fn}).

\section{Analysis}
After we empirically show the importance of selecting in-domain source data, the next question is: can we find a cost-effective way to nominate in-domain source data? 

\subsection{Measuring Similarity}
\label{section:measure-similarity}
We use three measures of the similarity between source and target data. We then observe whether these similarity values correlate with the usefulness of pretrained models in $\S$~\ref{section:correlation-analysis}.

\paragraph{Language model perplexity (PPL)}
has been used to provide a proxy to estimate corpus similarity~\citep{Baldwin:Cook:IJCNLP:2013}. We construct Kneser-Ney smoothed 3-gram models~\citep{Heafield:SMT:2011} on source data and use the perplexity of target data relative to these language models as the similarity between source and target data.

\paragraph{Jensen-Shannon divergence (JSD),} 
based on term distributions, has been successfully used for domain adaptation~\citep{Ruder:Plank:EMNLP:2017}. We first measure the probability of each term (up to 3-gram) in source and target data, separately. Then, we use the Jensen-Shannon divergence between these two probability distributions as the similarity between source and target data.

\paragraph{Target vocabulary covered (TVC)}
measures the percentage of the target vocabulary present in the source data, where only content words (nouns, verbs, adjectives) are counted. \citet{Dai:Karimi:NAACL:2019} show that it is very informative in predicting the effectiveness of pretrained word vectors.

\begin{figure}
    \centering
    \includegraphics[width=0.25\textwidth]{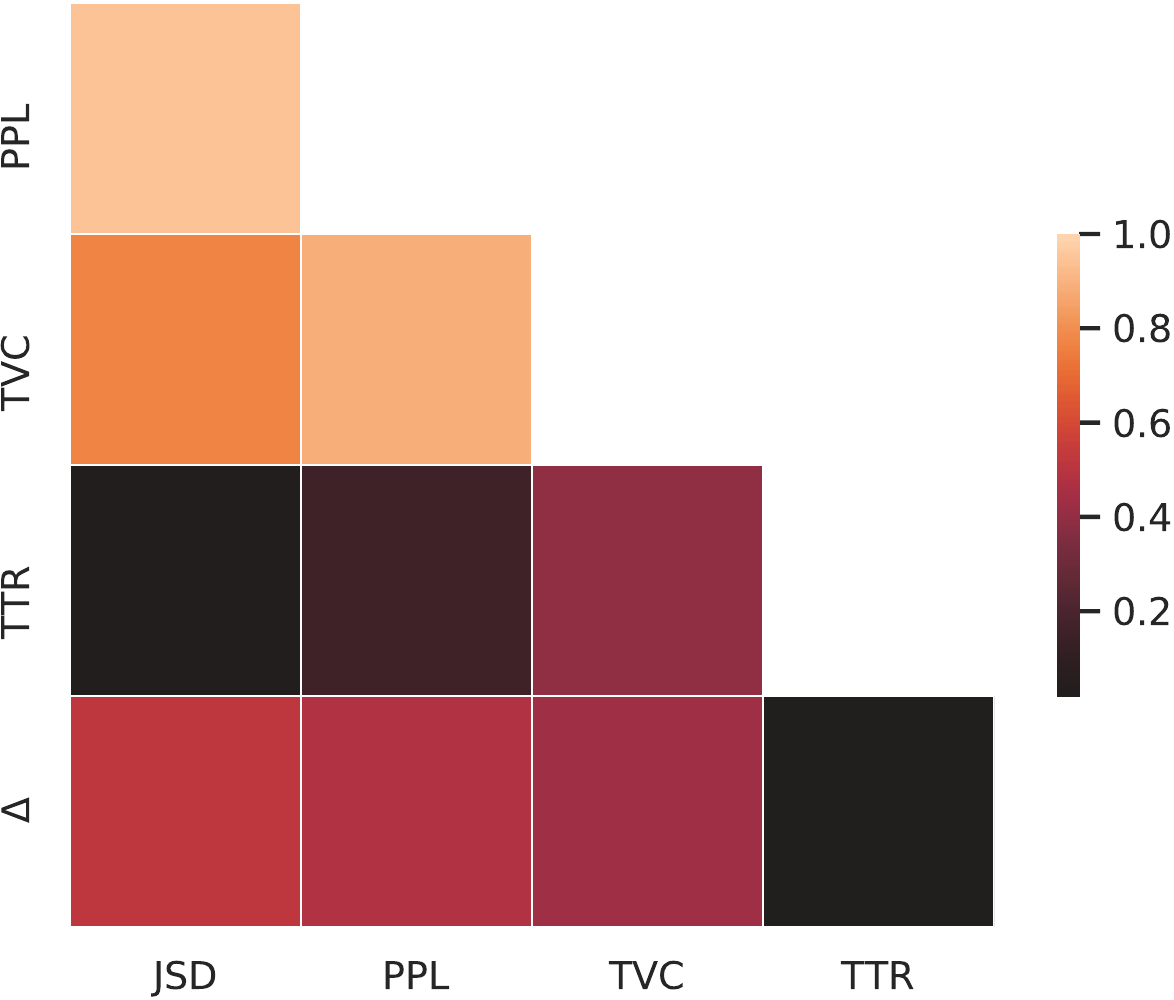}
    \caption{Correlation between different similarity measures and diversity measure and the improvement ($\Delta$) due to domain-specific BERT models.~\label{figure-correlation}}
\end{figure}

In addition, \citet{Ruder:Plank:EMNLP:2017} show that the diversity of source data is as important as domain similarity for domain adaptation. Inspired by this, we also explore a very simple diversity measure: type token ratio (\textbf{TTR}, $ \frac{\# \, \text{unique} \, \text{tokens}}{\# \, \text{tokens}}$),
that measures the lexical diversity of the source data.

To mitigate the impact of source data size on these measurements, for each source data, we sample five sub-corpora, each of which contains 10M tokens. Then we measure the similarity of source and target data and the diversity of source data as the average values of these sub-corpora.

\subsection{Correlation Analysis}
\label{section:correlation-analysis}
To analyze how the effectiveness of domain-specific BERT models correlate to the similarity between source and target data, we employ the Pearson correlation analysis to find out the relationships between improvements due to domain-specific BERT models and similarity between source and target data. For example, considering the BTC task, we use the performance of the original BERT as baseline, and measure the improvement due to Twitter BERT as $1.0$, whereas the corresponding value using BioBERT is $-2.9$. Note that we repeat all the experiments five times; therefore, we collect 300 source-target data points in total.

The correlation results are visualized in Figure~\ref{figure-correlation}. JSD has the strongest correlation (0.519) with the improvement due to domain-specific models, while the other two measures also have modest correlation ($0.481$ for PPL and $0.436$ for TVC). Recall that the calculation of JSD takes uni-grams, bi-grams and tri-grams into consideration, whereas PPL considers tri-grams only and the TVC considers uni-grams only. 
Correlations between different measures indicate that these measures are able to reach agreement on whether source and target are similar.
We find no correlation between the TTR of source data and the improvement.



\section{Summary}
We conduct a case study of pretraining BERT on social media text.
Through extensive experiments, we show the importance of selecting in-domain source data.
Based on empirical analysis, we recommend measures to help select pretraining data for best performance on new applications.

\section*{Acknowledgments}
We would like to thank anonymous reviewers for their helpful comments.
XD also thanks Shubin Du and Ying Zhou for early investigation of this work.
XD is supported by Sydney University's Engineering and Information Technologies Research Scholarship and a CSIRO Data61 top up scholarship. 

\bibliography{emnlp2020}
\bibliographystyle{acl_natbib}

\end{document}